\PassOptionsToPackage{table}{xcolor}
\documentclass[a4paper,fleqn]{cas-dc}

\usepackage{subcaption}
\usepackage{algorithm}
\usepackage{algorithmic}
\usepackage{xspace}
\newcommand{\ours}{VANDERER\xspace}
\usepackage[numbers,square,sort&compress]{natbib}

\usepackage{pifont}
\newcommand{\cmark}{\ding{51}}
\newcommand{\xmark}{\ding{55}}

\RenewDocumentCommand\printorcid{}{}

\begin{document}
\let\WriteBookmarks\relax
\def\floatpagepagefraction{1}
\def\textpagefraction{.001}

\shorttitle{VANDERER}    

\shortauthors{V.N. Devarakonda, R.G. Goswami, P. Krishnamurthy, F. Khorrami}  

\title [mode = title]{VANDERER: Map-Free Exploration using Future-Aware and Visual-Curiosity-Guided Diffusion Policy}

\author[1]{Venkata Naren Devarakonda}
\cormark[1]

\ead{d.naren@nyu.edu}

\credit{Conceptualization, Data curation, Methodology, Software, Writing – original draft}

\author[1]{Raktim Gautam Goswami}%
\ead{rgg9769@nyu.edu}
\credit{Methodology, Visualization, Writing – review and editing}

\author[1]{Prashanth Krishnamurthy}
\ead{prashanth.krishnamurthy@nyu.edu}
\credit{Supervision, Writing – review and editing}

\author[1]{Farshad Khorrami}
\ead{khorrami@nyu.edu}
\credit{Resources, Supervision, Project administration}

\affiliation[1]{organization={Control/Robotics Research Laboratory (CRRL), Department of Electrical and Computer Engineering, Tandon School of Engineering, New York University},
            addressline={5 Metrotech Center}, 
            city={Brooklyn},
            postcode={11201}, 
            state={NY},
            country={USA}}

\cortext[1]{Corresponding author}

\begin{abstract}
Mobile agents require efficient exploration strategies to map unseen environments and autonomously plan tasks. Traditional methods rely on generating occupancy maps and optimizing the sequence in which unexplored regions are visited. However, beyond requiring additional sensor suites for reliable pose and depth estimation, occupancy map-based exploration policies often neglect environmental traversability constraints. To address these limitations, we propose \ours, an exploration framework that leverages a Visual Curiosity Module (VCM) to guide pre-trained diffusion policies using only monocular image data. This curiosity module predicts the outcomes of proposed actions via a navigation world model and evaluates them through a curiosity cost. The cost then guides the diffusion process toward generating actions that maximize exploration. Evaluated across diverse simulated environments, \ours outperforms established baselines, achieving $2.3\times$ the exploration efficiency of Frontier Search. Our results reveal a correlation between visual and geometric curiosity in outdoor environments, demonstrating that \ours can effectively leverage this relationship for efficient exploration using simple robotic setups. The project video is available at \url{https://youtu.be/u1z1SVjK4_A}.
\end{abstract}

\begin{keywords}
 Autonomous Robot Exploration \sep Diffusion Policy \sep Diffusion Guidance \sep World Model
\end{keywords}

\maketitle

\section{INTRODUCTION}\label{sec:intro}
Autonomous exploration is a long-standing problem in robotics, with critical applications in search-and-rescue, environmental monitoring, and infrastructure inspection. Traditional frameworks rely on constructing global occupancy maps, necessitating precise localization and multi-sensor suites (GPS, IMUs, depth cameras, or LiDAR). Beyond the physical constraints, relying on such multi-sensor stacks introduces substantial calibration overhead, sensor-fusion complexity, and data collection bottlenecks that limit scalability across diverse robotic platforms. Additionally, relying purely on geometric occupancy mapping for exploration neglects traversability constraints posed by specific environments. While frontier-based strategies~\cite{Yamauchi1997AFA} are effective in structured indoor settings, outdoor environments introduce unique complications. For instance, geometric gaps between buildings may be flagged as frontiers even when physically impassable. Moreover, outdoor exploration is often governed by semantic or legal constraints, such as traffic regulations, which restrict movement to specific zones regardless of geometric accessibility. To address these limitations, this work explores how autonomous exploration can be conducted efficiently using pretrained diffusion policies that forego explicit geometric map constructions and simplify the sensor setup for real-world deployment and scalability.

Recent works in exploration and navigation have increasingly centered on learning-based approaches. A significant portion of such research relies on Reinforcement Learning (RL) to train task-specific policies~\cite{survey_garaffa}. However, RL-based methods typically require extensive training within individual environments and demand large amounts of high-quality data \citep{belkhale2023data}. Consequently, diffusion policies~\citep{chi2023diffusionpolicy} have emerged as a promising alternative, often reducing the training time required to adapt policies to new environments. Recent results demonstrate their success in tasks such as goal-directed navigation~\citep{sridhar2024nomad}, exploration~\citep{cao_dare_2025}, and collision-free planning~\citep{zeng2025navidiffusor}. Furthermore, diffusion policies have shown potential for generalizing across multiple tasks through intelligent guidance and masking strategies~\citep{sridhar2024nomad}. 

Environment exploration, however, presents a unique challenge. Unlike goal-directed navigation, where training trajectories are abundant, expert data for exploration is limited. In fact, exploration is not a strictly defined objective with a single optimal behavior. Rather, it is a broad goal that admits many valid trajectories. This raises a fundamental question: how can large-capacity models be trained to behave exploratively without relying on explicit expert exploration data? Additionally, how can existing datasets or scalable data generation be used to support such training?

\begin{figure*}
    \centering
       \includegraphics[width=1.0\textwidth]{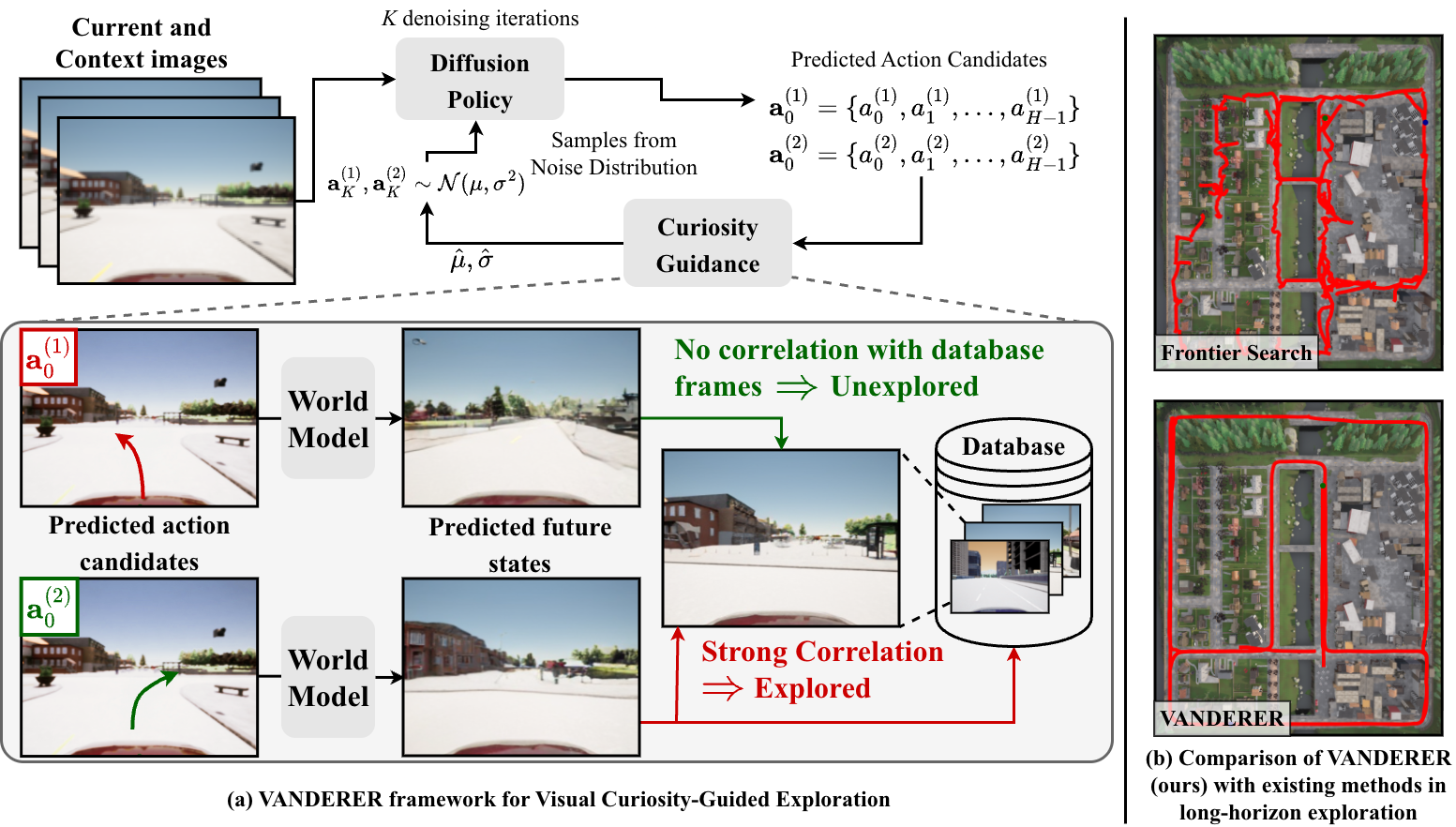}
       \caption{\textbf{VANDERER framework for guiding diffusion policies towards efficient autonomous exploration.} (a) A simplified example of action predictions generated from two noise latents sampled from a time-invariant Gaussian distribution through a pretrained diffusion policy. The curiosity guidance mechanism predicts the resulting future states and estimates their novelty against a database of prior observations. These novelty scores are then used to update the diffusion policy's noise distribution, guiding the agent toward unexplored areas. (b) In long-horizon autonomous exploration, \ours enables the agent to explore greater area with a higher efficiency than mapping-based methods such as Frontier Search.}
       \label{fig:intro}
\end{figure*}

To address these challenges, we propose \ours (Fig.~\ref{fig:intro}), a framework designed for efficient outdoor exploration that foregoes the overhead of complex mapping. Our approach leverages a diffusion policy to generate feasible candidate low-level actions directly from RGB observations.
Central to our method is the Visual Curiosity Module (VCM), which utilizes a navigation world model~\citep{bar2025navigation} to predict future states resulting from candidate actions. The VCM quantifies the ``exploratory value'' of these predicted states by comparing them against a database of prior observations via a curiosity-based cost function. By assigning higher costs to novel states, the VCM acts as diffusion guidance, steering the sampling process toward exploratory trajectory generation during inference.
Crucially, \ours eliminates the need for specialized exploration data. We evaluate our method through comparisons with competitive baselines and ablative studies. In summary, our key contributions are:
\begin{itemize}
    \item A novel end-to-end framework for map-free exploration using a visual-curiosity-guided diffusion policy.
    \item A visual curiosity module that predicts the consequences of actions and guides the diffusion policy to generate outputs with greater exploratory potential.
    \item An inference-stage optimization strategy that adapts pre-trained diffusion policies for efficient exploration without the need for additional training.
    \item Extensive experimentation on completely unseen environments showing \ours's superior exploration efficiency over existing methods.
\end{itemize}

\section{RELATED WORKS}\label{sec:related}

\subsection{Mapping-based Exploration}
\begin{figure}
    \centering
    \includegraphics[width=0.8\linewidth]{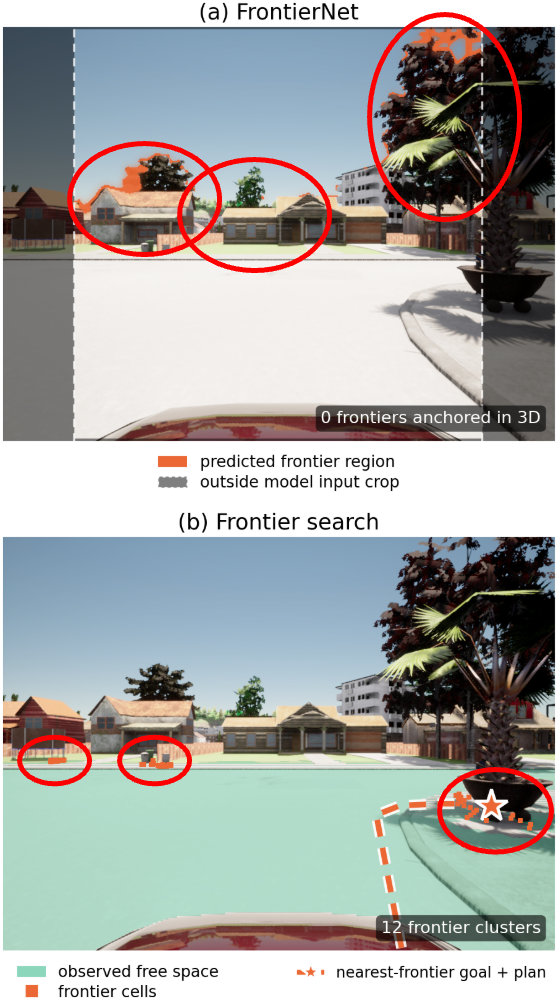}
    \caption{a) learned frontier methods~\citep{sun2025frontiernet}, and b) traditional frontier search~\citep{Yamauchi1997AFA} are implemented in CARLA simulation environments with RGB, Depth, and ground truth pose data, they detect invalid frontiers that would lead to collisions. While (b) might occasionally detect valid frontiers, (a) does not even detect frontiers on the ground.}
    \label{fig:frontiers}
\end{figure}

Mapping-based exploration focuses on expanding observed space by identifying navigable points. Frontier search methods~\citep{Yamauchi1997AFA, heng2015efficient} target occupancy grid boundaries, while sampling-based approaches \citep{2017umari_rrtexplore, rhbev_bircher16, lindqvist2021exploration} explore unoccupied areas via tree or graph structures \citep{LaValle1998RapidlyexploringRT, karaman2011sampling}. Hybrid frameworks \citep{hybrid_selin19} attempt to balance these by combining frontier guidance with sampling-based local exploration. Crucially, all these methods rely on accurate map building using multiple sensors, a hardware constraint our monocular RGB approach explicitly removes. Beyond that, as shown in Fig.~\ref{fig:frontiers}b, it is observed that mapping outdoor environments using depth or lidar sensors leads to detection of frontiers outside the agent's legally traversable region. With no obvious way to filter such frontiers, these methods can lead to unsafe exploration paths that result in collisions.

While newer methods have been developed to mitigate this requirement by leveraging trainable neural networks to use 2D visual cues to predict frontiers~\citep{sun2025frontiernet}, such approaches still have fundamental downsides. As shown in Fig.~\ref{fig:frontiers}a, some methods entirely fail to detect any feasible frontiers in unseen environments. Others do not scale well to larger environments, particularly outdoor ones, where the explorable environments might not be bounded, or might be too large to map efficiently (refer to the evaluations in Sec.~\ref{subsec:results}). On top of that, in more structured environments like autonomous driving scenarios with additional constraints such as traffic regulations, such methods might lead to infeasible exploration paths (as shown in Fig.~\ref{fig:frontiers}).

\subsection{RL-based approaches}
A prominent class of learning-based exploration approaches utilize RL frameworks to train policies through various intrinsically motivated reward structures, such as future-state prediction error in a learned latent space~\citep{pmlr-v70-pathak17a}. Curiosity can also be measured using network distillation that compares the output of a trained predictor with that of an untrained one, using the error as a signal for curiosity~\citep{DBLP:conf/iclr/BurdaESK19}. A more recent approach uses the error between the predictions of an ensemble of single-step dynamic models as curiosity reward~\citep{pathak2019self}. In contrast, several deep reinforcement learning methods \citep{cao2023ariadne} use extrinsic reward obtained from additional sensors such as depth cameras and LiDARs to abstract information like explorative frontiers and occupancy maps~\citep{li2019deep,chen2025gleam,2024cao_drl}. While these strategies are effective, the requirement for environment-specific training in simulation often limits their direct applicability to real-world exploration.
\subsection{Diffusion Policies}
Analogous to map-based extrinsic rewards in RL, diffusion policies can learn exploration by conditioning on map representations \citep{cao_dare_2025,tan20264cnet}. More specifically, recent works have utilized raw RGB frames from real-world trajectories to learn goal-based navigation \citep{shah_vint_2023,pmlr-v164-shah22a,yokozawa2025deep}. Diffusion policies have further enabled unified frameworks that learn both navigation and exploration within a single model \citep{sridhar2024nomad}, with subsequent improvements incorporating learned metric scaling \citep{nayak2025metricnet} and conditional flow matching \citep{gode2025flownav}. While these RGB-only methods perform well for goal-conditioned tasks, they rely primarily on policy variance for discovery, lacking the global information necessary for long-horizon exploration.

Another advantage of diffusion policies is their capacity for objective-specific guidance using pre-trained models. This can be achieved through trained classifiers that steer the denoising loop \citep{dhariwal2021diffusion} or via classifier-free guidance, which uses a weighted combination of class-dependent and class-agnostic generations \citep{ho2021classifier}. Recent research has also focused on optimizing the initial noise distribution for more targeted generation \citep{guo2024initno,chen2024find,karunratanakul2024optimizing}. In goal-based navigation, guidance techniques have already been applied to collision avoidance and safety \citep{pmlr-v305-ma25a,zeng2025navidiffusor}. Our work builds on these methods, proposing a curiosity-based guidance mechanism that optimizes diffusion outputs for long-range exploration using only a monocular RGB stream.

\begin{figure*}[t]
    \centering
    \includegraphics[width=1.0\linewidth]{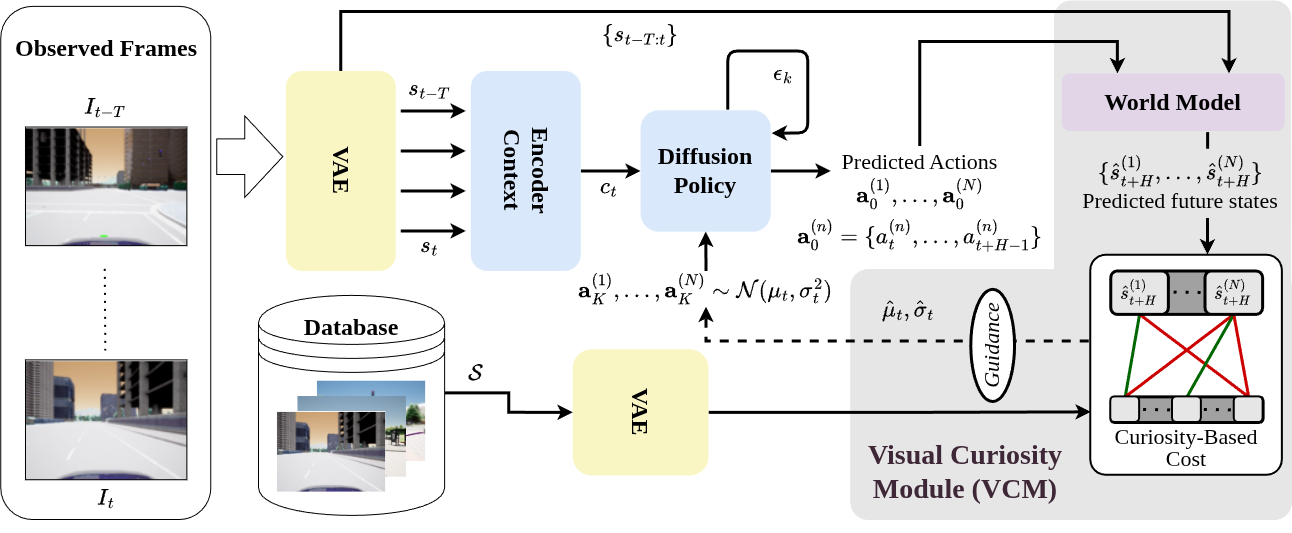}
    \caption{The current ($I_t$) and T prior frames ($I_{t-T:t-1}$) are separately encoded using a Variational Autoencoder (VAE). A context encoder generates a conditioning vector from the states for policy generation. The world model in the VCM autoregressively predicts sequences of future states using the observed frames and policy-generated action sequences. The curiosity-based cost function matches each predicted final state with the previously observed states in the database. The most novel candidate is used to update the diffusion noise distribution.}
    \label{fig:main}
\end{figure*}

\section{METHODOLOGY}\label{sec:method}

\subsection{Problem Statement}
We consider an agent equipped with a monocular camera traversing an unmapped, city-scale outdoor environment. At each discrete timestep $t$, the agent receives an RGB observation $I_t \in \mathbb{R}^{\mathcal{H} \times \mathcal{W} \times 3}$, ($\mathcal{H}$: height; $\mathcal{W}$: width of the image) which serves as the sole input for the waypoint predictions. The objective of \ours is to generate navigation commands that maximize exploration efficiency, which in the case of our CARLA environments, is defined as the ratio of unique road length explored to cumulative distance traveled. These waypoints are predicted in the 2D top-down projection of the agent's local coordinate frame and can be executed using a suitable low-level controller.

\subsection{Method Overview}
Existing diffusion policies for mobile agents \citep[e.g.,][]{sridhar2024nomad} typically focus on goal-conditioned navigation and use future frames in trajectory datasets as goals during training. However, the resulting paths lack global context and are rarely optimized for exploration. 
Further, scaling 2D occupancy grids, capable of generating exploratory trajectories, to large outdoor environments is difficult due to complex global mapping requirements. To address these limitations, we employ a ``generate-then-optimize'' framework. Our diffusion policy is trained to imitate trajectory data to predict feasible and safe action sequences. During inference, we guide the model toward exploration-friendly behaviors by optimizing the initial noise latent distribution within the reverse diffusion process.

At the core of \ours is the VCM (Sec.~\ref{sec:vcm}), which guides a Diffusion Policy (Sec.~\ref{sec:diff_policy}) to generate actions tailored for environmental exploration (Fig.~\ref{fig:main}). \ours first maps the current observation $I_t$ and $T$ historical frames to latent states $\{s_{t-T}, \dots, s_t\}$ using patch-level features from a frozen VAE ~\citep{blattmann2023stablevideodiffusionscaling,bar2025navigation}. 
These states are encoded into a conditioning vector $c_t$, allowing the diffusion policy to iterate from a large denoising timestep ($K$) to 0, eventually producing an action sequence $\{a_0, \dots, a_{H-1}\}$ representing relative $(x, y)$ coordinate changes. Sampling multiple noise latents, $N$ such action candidates are generated. In notation, a variable $x_t^{(n)}$ denotes the n-th sample of the variable $x$ at time $t$. Further, bold lettered actions $\mathbf{a}_k^{(n)}$ denotes the n-th action sequence sample generated at the k-th denoising timestep by the diffusion policy. Finally, the VCM processes these actions and states to predict future observations, which inform a latent guidance loss that optimizes the diffusion process for maximum environmental exploration.

\subsection{Visual Curiosity Guidance}
\label{sec:vcm}
The VCM predicts the consequences of the candidate actions generated by the diffusion policy to steer the agent toward exploration. This mechanism comprises three core components: future state prediction via a world model, a curiosity-based cost function, and the application of this loss to provide diffusion guidance.

\subsubsection{Predicting Future States}
The action sequence generated by the diffusion policy, combined with context and current latent states, is fed into a world model to predict (\textit{imagine}) future latent states $\{\hat{s}_{t+1}, \dots, \hat{s}_{t+H}\}$. We employ the Navigation World Model (NWM) framework~\citep{bar2025navigation}, which utilizes patch-level VAE features as states and defines the action vector as the relative change ($\Delta$) in $x$-coordinate, $y$-coordinate, and yaw. With $x$ and $y$ displacement values from $a_i$, we estimate the change in yaw as $\text{atan2}(\Delta y,\Delta x)$.

\begin{algorithm}
\caption{Curiosity Loss Computation}
\label{alg:curiosity_loss}
\begin{algorithmic}[1]
\renewcommand{\algorithmiccomment}[1]{\hfill // \textit{#1}}

\STATE \textbf{Input:} Predicted future states $\{\hat{s}_{t+H}^{(n)}\}_{n=1}^N$, Database $\mathcal{S}$
\STATE \textbf{Output:} Curiosity costs

\STATE $\textit{costs} \gets [\;]$
\FOR{each predicted future state $\hat{s}_{t+H}^{(n)}$}
    \STATE $\textit{distances} \gets [\;]$
    \FOR{each visited state $s \in \mathcal{S}$} 
        \STATE $\textit{dists} \gets \mathrm{fast\_recip\_nearest\_neighbors}(\hat{s}_{t+H}^{(n)}, s)$
        \STATE $d \gets \mathrm{mean}(\mathrm{topk}(\textit{dists}, 250))$
        \STATE \text{append } $d$ \text{ to } $\textit{distances}$
    \ENDFOR
    \STATE $\mathcal{L}_{c}^{(n)} \gets \min(\textit{distances})$
    \STATE \text{append } $\mathcal{L}_{c}^{(n)}$ \text{ to } $\textit{costs}$
\ENDFOR

\STATE \textbf{return} $\textit{costs}$
\end{algorithmic}
\end{algorithm}

\subsubsection{Curiosity-Based Cost Function}
Given the predicted future states, we want to quantify their exploratory value. In the absence of global maps or odometry, we assess novelty by comparing the final predicted state $\hat{s}_{t+H}$ against the database of previously observed states $\mathcal{S} = \{s_0, \dots, s_t\}$. Since these latent states correspond to visual observations, we hypothesize that in environments with sufficient visual variance, maximizing visual novelty effectively maximizes environmental coverage.  To compute the similarity $\mathcal{D}$ between $\hat{s}_{t+H}$ and a reference state $s_j \in \mathcal{S}$, we utilize MASt3R’s~\citep{leroy2024grounding} fast reciprocal matching on patch-level features (Algorithm~\ref{alg:curiosity_loss}). Specifically, an iterative reciprocal nearest neighbor search establishes a one-to-one mapping between patches in $\hat{s}_{t+H}$ and their closest counterparts in $s_j$. The similarity $\mathcal{D}$ is defined as the mean Euclidean distance between the \textit{top 250} most distant corresponding patch pairs. The final curiosity cost $\mathcal{L}_c$ is determined by the minimum distance across all states in $\mathcal{S}$, identifying the observation most similar to the predicted future state:
\begin{equation}
    \mathcal{L}_c(\hat{s}_{t+H}, \mathcal{S}) = \min_{s_j \in \mathcal{S}} \mathcal{D}(\hat{s}_{t+H}, s_j).
\end{equation}

This patch-matching approach is more robust than pixel-wise or index-based comparisons, where minor camera shifts can cause significant errors if indices no longer align with the same visual features. Additionally, visual similarity metrics typically saturate beyond a certain threshold, failing to reflect true spatial separation between distant, distinct locations. To improve efficiency and reduce computational load, we compare the predicted state $\hat{s}_{t+H}$ against a temporally subsampled subset of $\mathcal{S}$, thereby filtering out the redundant information found in adjacent frames.

\subsubsection{Diffusion Guidance}
\begin{figure}
    \centering
    \fbox{\includegraphics[width=0.96\linewidth]{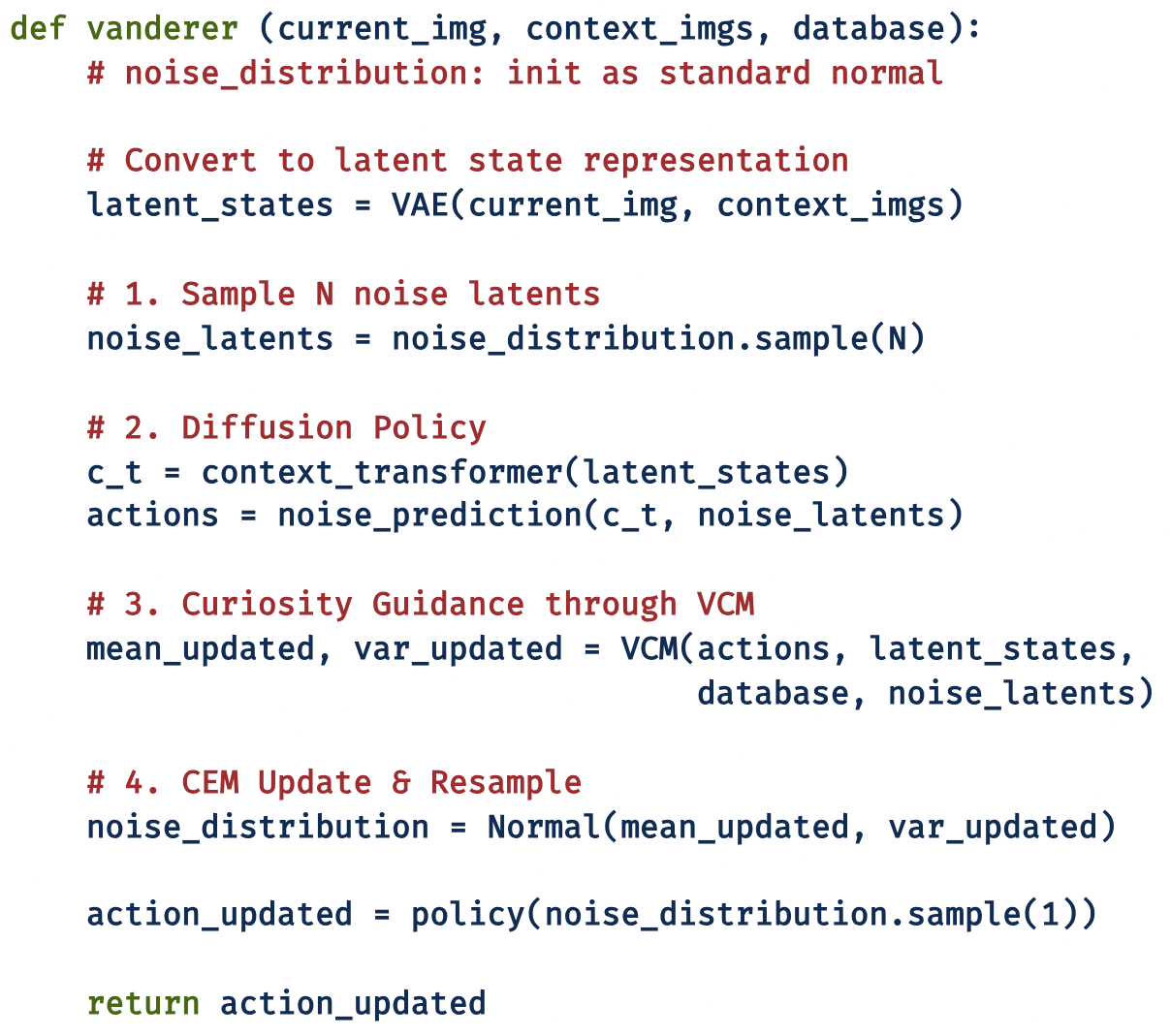}}
    \caption{\textbf{Pseudocode of a single-step action guidance} using VCM. The guidance algorithm initiates once the action candidates exhibit sufficient variance, at which point the noise distribution is updated based on VCM scores. Finally, the optimized action is reconstructed by the diffusion policy model using this refined noise distribution.}
    \label{fig:pseudocode}
\end{figure}

The curiosity cost $\mathcal{L}_c$ guides the diffusion process toward actions that prioritize exploration (Fig.~\ref{fig:pseudocode}). Specifically, $\mathcal{L}_c$ is used to refine the mean and variance of the diffusion policy's noise input, as different noise latents yield distinct, feasible action sequences. Although the noise could be updated via various optimization strategies, we find that sampling-based methods like the Cross-Entropy Method (CEM) yield superior results, consistent with recent world model literature~\citep{bar2025navigation,dexwm}.  The policy initially draws $N$ noise samples from a standard Gaussian distribution ($\mathcal{N}(\mu_0 = 0, \sigma_0 = I_{2H})$), generating $N$ candidate action sequences that are subsequently evaluated via $\mathcal{L}_c$. From these, the top-$r$ performing candidates are selected to empirically calculate the updated mean $\mu_1$ of the noise distribution. This process continues for $L$ iterations. Finally, the noise sampled from this updated distribution is fed to the policy to generate the optimal action sequence. To maintain generation stability, the noise distribution is reset to a standard Gaussian every $P$ execution steps. Notably, guidance is only applied when there is sufficient variance among the generated action candidates to ensure effective steering.

\begin{table*}[!htbp]
\centering
\small
\setlength{\tabcolsep}{2.7pt}
\caption{Mean performance metrics across five \textbf{unseen} test environments (Towns 1-5) arranged in increasing order of size. RC: Road Coverage ($\%$), REE: Road Exploration Efficiency ($m/m$), Avg PF: Average Policy Failure across towns. Each town is denoted by its name, followed by the total drivable road length in parenthesis. \ours beats all baselines in the overall exploration metrics, exhibiting higher gains as the size of the environment increases. Exploration in smaller towns leads to higher road loop repetitions due to the limited area to explore and limited road intersections. The presented standard deviation is a measure of how much the runs for any town deviate from their empirical mean. Hence, it is computed from the variances across the three seeds within each town and then pooled over the five towns.}
\label{tab:naviexp_results}
\begin{tabular}{l cc cc cc cc cc | ccc}
\toprule
& \multicolumn{4}{c}{\textbf{Small Towns}} & \multicolumn{6}{c|}{\textbf{Large Towns}} & \multicolumn{3}{c}{\textbf{Overall Average}} \\
\cmidrule(lr){2-5} \cmidrule(lr){6-11} \cmidrule(lr){12-14}
& \multicolumn{2}{c}{\shortstack{\textbf{Town 2} \\ (1,209 m)}} & \multicolumn{2}{c}{\shortstack{\textbf{Town 1} \\ (2,732 m)}} & \multicolumn{2}{c}{\shortstack{\textbf{Town 3} \\ (3,402 m)}} & \multicolumn{2}{c}{\shortstack{\textbf{Town 5} \\ (4,401 m)}} & \multicolumn{2}{c|}{\shortstack{\textbf{Town 4} \\ (6,370 m)}} & \multicolumn{2}{c}{} & \\
\cmidrule(lr){2-3} \cmidrule(lr){4-5} \cmidrule(lr){6-7} \cmidrule(lr){8-9} \cmidrule(lr){10-11} \cmidrule(lr){12-13} \cmidrule(lr){14-14}
Method & RC & REE & RC & REE & RC & REE & RC & REE & RC & REE & RC & REE & Avg PF \\
\midrule
Frontier Exp.~\citep{Yamauchi1997AFA} & 61.8 & 0.20 & 31.5 & 0.39 & 24.1 & 0.24 & 7.9 & 0.09 & 13.8 & 0.22 & 27.8 $\pm$ 16.0 & 0.23 $\pm$ 0.11 & 2.14\% \\
NBVP~\citep{rhbev_bircher16} & 69.3 & 0.21 & 29.8 & 0.29 & 27.1 & 0.24 & 28.8 & 0.32 & 21.6 & 0.35 & 35.3 $\pm$ 9.0 & 0.28 $\pm$ 0.04 & 3.85\% \\
NoMaD \citep{sridhar2024nomad} & \cellcolor{gray!20}\textbf{82.9} & \cellcolor{gray!20}\textbf{0.25} & 59.0 & 0.41 & 44.0 & 0.38 & 44.3 & 0.49 & 38.4 & 0.62 & 53.7 $\pm$ 12.4 & 0.43 $\pm$ 0.09 & \cellcolor{gray!20}\textbf{0.14\%} \\
DP-RND \citep{DBLP:conf/iclr/BurdaESK19} & 64.0 & 0.20 & 70.0 & 0.49 & 45.7 & 0.40 & 39.6 & 0.44 & 45.6 & 0.74 & 53.0 $\pm$ 5.3 & 0.45 $\pm$ 0.05 & 0.41\% \\
\textbf{VANDERER} & 67.7 & 0.21 & \cellcolor{gray!20}\textbf{78.1} & \cellcolor{gray!20}\textbf{0.55} & \cellcolor{gray!20}\textbf{49.2} & \cellcolor{gray!20}\textbf{0.46} & \cellcolor{gray!20}\textbf{59.4} & \cellcolor{gray!20}\textbf{0.67} & \cellcolor{gray!20}\textbf{50.5} & \cellcolor{gray!20}\textbf{0.82} & \cellcolor{gray!20}\textbf{61.0 $\pm$ 6.8} & \cellcolor{gray!20}\textbf{0.54 $\pm$ 0.07} & 0.29\% \\
\bottomrule
\end{tabular}%
\end{table*}

\begin{table}[!htpb]
    \centering
    \caption{Input modalities required by each method.}
    \label{tab:naviexp_modalities}
    \begin{tabular}{l cccc}
    \toprule
    Method & RGB & Global Pose & Depth & Avg RC\\
    \midrule
    Frontier Exp. & \xmark & \cmark & \cmark & 27.8\%\\
    NBVP                                 & \xmark & \cmark & \cmark & 35.3\%\\
    NoMaD        & \cmark & \xmark & \xmark & 53.7\%\\
    DP-RND & \cmark & \xmark & \xmark & 53.0\%\\
    \textbf{VANDERER}                    & \cmark & \xmark & \xmark & \cellcolor{gray!20}\textbf{61.0\%}\\
    \bottomrule
    \end{tabular}
\end{table}

\subsection{Diffusion Policy}
\label{sec:diff_policy}
To employ the diffusion policy, all frames are first processed by a VAE encoder into 3D feature tokens, which are then positionally encoded to preserve spatial and temporal properties. These tokens are processed by a context transformer, comprising six standard transformer blocks with 8 attention heads in a 256-dimensional space, before being pooled over the spatial and temporal dimensions into a 1D vector ($c_t$) to serve as conditioning for the diffusion model. 

This model utilizes a UNet-style architecture. During training, following the DDPM framework, Gaussian noise is added to ground-truth action sequences according to Eq.~\ref{eq:noisy input} using a variance schedule for $K$ diffusion timesteps, defined by parameters ($\beta_1,\dots,\beta_K$) and the following constants:
\begin{gather}
    \alpha_k = 1 - \beta_k, \; \; \; \; \bar{\alpha}_k = \prod_{i=1}^k \alpha_i, \; \; \; \; \tilde{\beta}_k = \frac{1-\bar{\alpha}_{k-1}}{1-\bar{\alpha}_{k}}\beta_k \\
    \epsilon \sim \mathcal{N}(0, I) \;\;\;\; k \in \{1,\dots,K\}.
\end{gather}
Following Eq.\ref{eq:noisy input}, noisy inputs ($\mathbf{a}_k$) are generated from ground truth actions. The noise prediction model ($\epsilon_{\theta}(\mathbf{a}_k, k)$) is optimized to estimate this noise, given the noisy input:
\begin{align}
    \mathbf{a}_k &= \sqrt{\bar{\alpha}_k} \mathbf{a}_0 + \sqrt{1 - \bar{\alpha}_k} \epsilon \label{eq:noisy input} \\
    \mathcal{L}_{diff} &= \mathbb{E}_{k, \mathbf{a}_0, \epsilon} \left[ \| \epsilon - \epsilon_\theta(\mathbf{a}_k, k, c_t) \|^2 \right].
    \label{eq:diff loss}
\end{align}
During inference, the action prediction at any rollout timestep $t$ begins by initializing the noisy input ($\mathbf{a}_{K} \sim \mathcal{N}(0, I)$), which is iteratively refined to obtain the denoised action sequence $\mathbf{a}_{0}$. At denoising timestep $k$, the estimated noise $\epsilon_{\theta}(\mathbf{a}_k, k)$ is used to predict the clean action ($\hat{\mathbf{a}}_0$), which in turn is used to compute ${\mathbf{a}}_{k-1}$. This process continues until $k=1$ for $K=10$ timesteps, to derive the final denoised output using the following equations: 
\begin{align}
    \hat{\mathbf{a}}_0 &= \frac{\mathbf{a}_k - \sqrt{1 - \bar{\alpha}_k} \epsilon_\theta(\mathbf{a}_k, k, c_t)}{\sqrt{\bar{\alpha}_k}} \\
    \mathbf{a}_{k-1} \sim \mathcal{N}&\left(\tilde{\mu}_k(\mathbf{a}_k,\hat{\mathbf{a}}_0), \tilde{\beta}_k I\right) \;\;\;\; \forall k \in \{K,\dots, 1\} \\
    \tilde{\mu }_{k}(\mathbf{a}_{k},\hat{\mathbf{a}}_{0}) &= \frac{\sqrt{\bar{\alpha}_{k-1}}\beta_k}{1-\bar{\alpha}_k}\hat{\mathbf{a}}_0 + \frac{\sqrt {\alpha _k}(1-\bar{\alpha}_{k-1})}{1-\bar{\alpha}_k}\mathbf{a}_k.
\end{align}

\section{EXPERIMENTS}
Our experiments are designed to address the following key questions: (a) How effectively does \ours navigate and explore large-scale outdoor environments?  (b) What is the impact of curiosity-based diffusion guidance on overall performance?  (c) How does the explored area scale with respect to experiment duration?  (d) Which components of \ours are most critical to its success?

\subsection{Experiment Settings}
\vspace{1.5mm}
\noindent\textbf{Dataset.} We evaluate \ours using the CARLA~\citep{pmlr-v78-dosovitskiy17a} simulator, which provides a robust testing suite of multiple outdoor environments featuring varied scales and diverse visual features. In our setup, the agent is a vehicle equipped with an ego monocular camera. To simplify the environments, we removed lamp posts from the scenes. We use five towns (Towns 6, 7, 10, 11, and 12) for finetuning our models, while evaluation is conducted on five previously unseen towns (Towns 1--5). The training data consists of collision-free trajectories generated via CARLA’s \textit{BasicAgent} between random waypoints within a set distance threshold. Since some training towns contain an unequal proportion of straight-line roads to intersections, we adjust the number of trajectories collected at turns to balance the dataset and improve the fine-tuned policy's performance. To avoid overfitting to the training towns, the position of the sun and the weather conditions are varied across different trajectories. We evaluate each environment across three random seeds to account for the stochastic nature of the learned models.

\vspace{1.5mm}
\noindent\textbf{Training Settings.} We train our models using two NVIDIA L40S GPUs. The Diffusion Policy (Sec.~\ref{sec:diff_policy}) is first pretrained on the TartanGround dataset~\citep{patel2025tartanground} (comprising 878 trajectories across 63 photorealistic environments, totaling 1.44M samples) for 100 epochs. Freezing the context transformer's weights, the noise prediction part is subsequently fine-tuned for 30 epochs on the previously mentioned collective CARLA training dataset. All learning-based baselines are finetuned from their publicly released checkpoints on the exact same dataset for the same number of epochs to ensure a fair comparison. For the Visual Curiosity Module (Sec.~\ref{sec:vcm}), we fine-tune the ``base'' variant of the Navigation World Model~\citep{bar2025navigation} from its open-sourced pretrained checkpoint on the same CARLA training dataset for 25 epochs. During evaluation, proposed waypoints are executed via a low-level PID controller, with replanning triggered every 40 simulation steps ($\sim$2~seconds).

\vspace{1.5mm}
\noindent\textbf{Inference.} We perform all evaluations on a device with a single NVIDIA RTX 4070 GPU. The two critical models involved in \ours govern its overall planning latency. The action generating diffusion policy is sampled using DDPM with 10 denoising timesteps. A total of 20 candidate actions are generated at each planning step, followed by a farthest point sampling process that identifies the 6 most varying action proposals for guidance by the VCM. 

Since the world model is used to predict the most probable outcome of a candidate trajectory rather than to generate a diverse set of possible futures, we replace the stochastic DDPM sampler with a deterministic DDIM sampler. This change only affects the denoising sampling procedure and DDIM is applied directly to the model weights obtained from DDPM-based finetuning. DDIM defines a non-Markovian trajectory that remains valid over arbitrary subsequences of timesteps, allowing us to use only 20 denoising steps for the future state generation. Despite 7.5× fewer function evaluations as compared to the DDPM sampling with 150 denoising steps, DDIM with 20 denoising steps matches and in some cases improves the generated output latents. The world model takes a total of 0.84 s to generate the final state prediction for all the 6 filtered action candidates. This enables a real-time planning frequency of $\sim1.2$ Hz. However, as stated previously, all evaluations run policies at a frequency of $\sim0.5$ Hz.

\begin{figure*}
    \centering
    \includegraphics[width=1\linewidth]{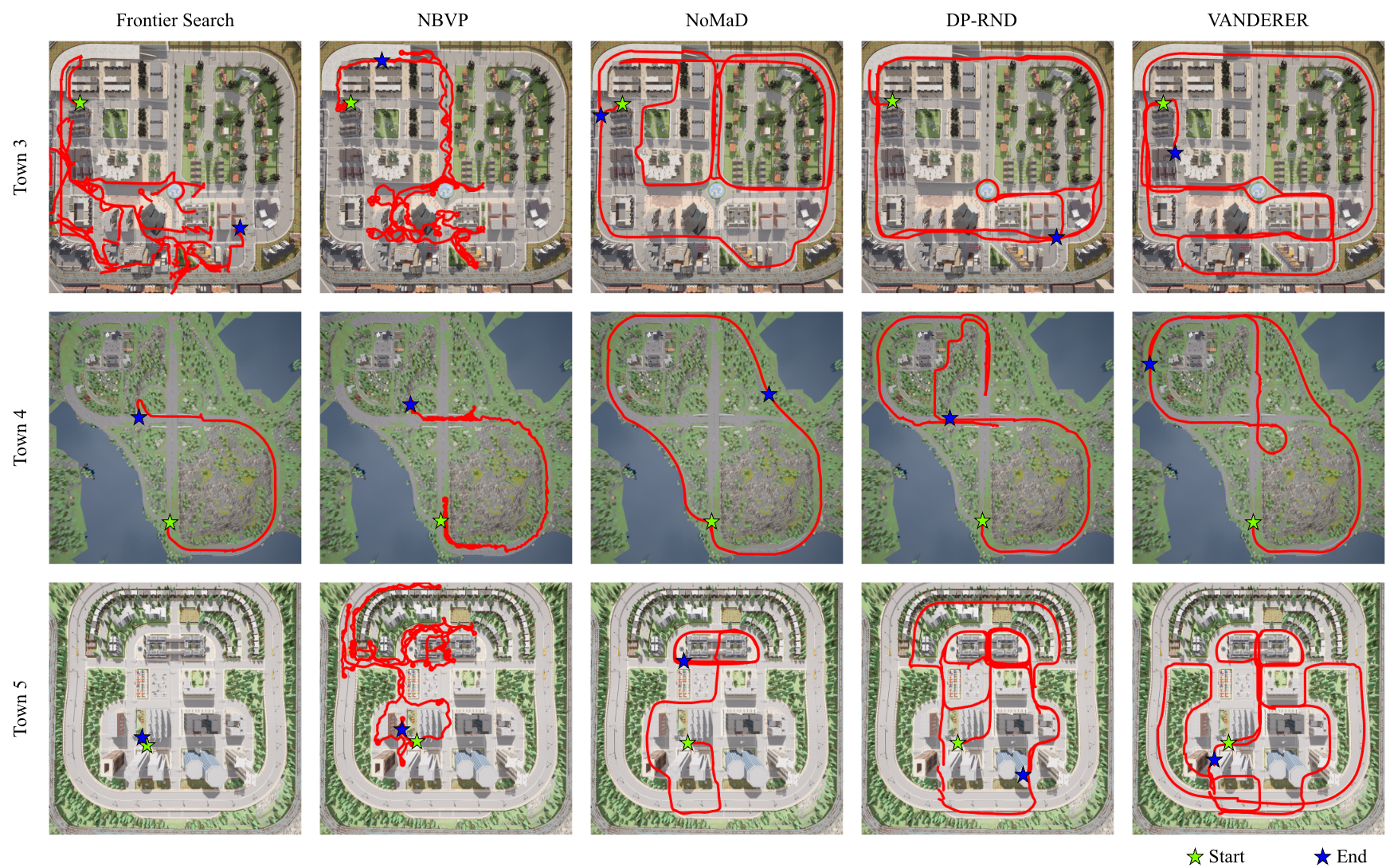}
    \caption{Qualitative comparison of resultant trajectories across three environments (Town 3, 4, and 5). The red lines illustrate the paths taken by each agent from the \textbf{start (green star)} to the \textbf{end (blue star)}. The map-based methods like Frontier Search and NBVP often choose frontiers/views that lead to obstacles and less efficient exploration. The diffusion policy-based baselines like NoMaD and DP-RND often exhibit path repetitions. It is evident that \ours demonstrates higher efficiency in exploration. }
    \label{fig:exploration_qualitative}
\end{figure*}

\subsection{Baselines} \label{subsec:baselines}
We evaluate the exploration performance of \ours against the following state-of-the-art methods:  

\begin{itemize}
    \item \textbf{NoMaD}~\citep{sridhar2024nomad}: An RGB-only exploration method using a diffusion policy trained on a large-scale navigation dataset. Hence, we fine-tuned the NoMaD model on our CARLA dataset for the same number of epochs.
    \item \textbf{DP-RND} \citep{DBLP:conf/iclr/BurdaESK19}: Random Network Distillation (RND) is an intrinsic reward used in RL training to explore unseen states. We define a baseline using RND as a selection metric for our diffusion policy to compare against our VCM.
    \item \textbf{Frontier Search} \citep{Yamauchi1997AFA}: We use Yamauchi's nearest-frontier method with explore\_lite's\footnote{https://github.com/robo-friends/m-explore-ros2} goal-selection conventions. Because it needs depth and localization, we build its occupancy grid from a depth camera at the RGB camera's pose, using ground-truth poses from the simulator. Paths are executed by the same PID controller as the other methods.
    \item \textbf{Next Best View Planner (NBVP)}~\citep{rhbev_bircher16}: We adapt its official open-source implementation into our CARLA testing setup. Similar to Frontier Search, NBVP also requires the generation of a 2D metric occupancy map, which is achieved through additional depth camera and ground-truth pose data from the simulator. Further, the viewpoint sampling is adapted to work with the vehicle agent in our CARLA setup.
\end{itemize}

Table~\ref{tab:naviexp_modalities} highlights the input modalities required by \ours and the baselines for exploration. The ego vehicle observes the world through a single forward-facing RGB camera ($320\!\times\!240$, $90^\circ$ FOV) mounted on it, providing the visual input for the diffusion policy-based methods (NoMaD, DP-RND, and \ours). The target waypoints provided by each method are executed using a low-level PID controller which tracks at a target speed of $3.6$\,km/h. Instead of the RGB frames, the geometric planning baselines (Frontier Search and NBVP) require metric depth and global positioning to generate a world-frame 2D occupancy grid. The diffusion policy-based methods do not need global positioning as their predicted waypoints are expressed in the agent's own frame. Hence, local ego odometry is sufficient to drive the PID controller. For ease and uniformity of evaluation, however, all methods are supplied with ground-truth positioning from the simulator, so localization error is removed as a confound and the comparison isolates exploration behaviour. A collision sensor is attached purely to detect collisions and respawn the vehicle to the nearest drivable spawn point to resume exploration.

\subsection{Evaluation}
\label{subsec:evaluations}
\vspace{1.5mm}
\noindent \textbf{Evaluation Setup.}
We evaluate on CARLA Towns 1--5, which feature diverse layouts, characteristics, and scales. This ranges from a small suburb to a multi-junction urban map. Each method is run with three random seeds per town ($15$ runs per method) under an identical budget of $100{,}000$ synchronous simulation steps at $\Delta t\!=\!0.05$\,s. So every method receives the same number of policy calls and the same amount of driving time.

\vspace{1.5mm}
\noindent \textbf{Evaluation Metrics.}
Autonomous exploration in bounded indoor environments is traditionally evaluated using total coverage area (cumulative free-space observed within a fixed step budget) or time-to-coverage (the time required to achieve total coverage). In outdoor autonomous driving, however, traversable space is constrained to a sparse road network. Area-based metrics like total coverage area falsely reward illegal off-road driving, and time-to-coverage is hard to compute due to the large scale of outdoor environments. Hence, we evaluate performance directly along the underlying road network using two primary metrics:
\begin{itemize}
    \item Road Coverage (RC): The fraction of the town's total road network length visited by the agent, discarding any off-road exploration paths.
    \item Road Exploration Efficiency (REE): The unique road length explored divided by the total distance driven. REE is bounded by 1.0 (achieved only when never re-traversing road segments) and penalizes backtracking or localized circling.
\end{itemize}

\subsection{Results}
\label{subsec:results}
\vspace{1.5mm}
\noindent\textbf{Exploration Performance.}
A quantitative comparison of \ours against the baselines is presented in Table~\ref{tab:naviexp_results}. Leveraging curiosity-guidance, \ours consistently achieves superior exploration performance. The presented values show an interesting trend of a larger performance gap between \ours and other methods as the size of the environment increases. The majority of the road network in Town 4 is part of a large ringroad, with limited exits. This makes it the easiest town to explore, as indicated by the relatively high REE values. The other larger towns like 3 and 5 also show a clear difference in both coverage and efficiency between \ours and the other methods.

The effectiveness of our method is evident even in smaller environments. In Town 1, \ours attains both the highest coverage ($78.1\%$) and the highest REE ($0.55$). 
Even in Town 2, the smallest test map where a single rectangular loop dominates two-thirds of the road network, our guidance mechanism yields a $14.7\%$ relative improvement over the unguided base policy.

\begin{figure}[]
    \centering
    \begin{subfigure}[b]{0.48\linewidth}
        \centering
        \includegraphics[width=\linewidth]{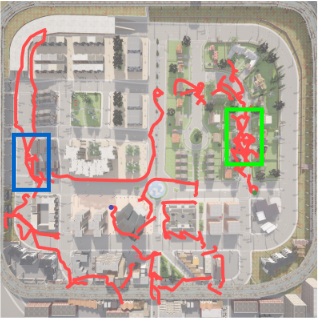}
        \caption{Frontier Search}
    \end{subfigure}
    \hfill
    \begin{subfigure}[b]{0.48\linewidth}
        \centering
        \includegraphics[width=\linewidth]{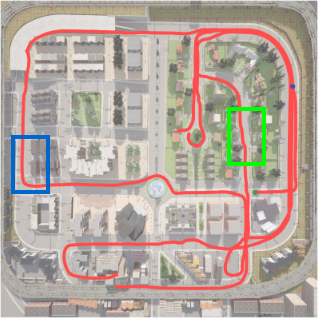}
        \caption{VANDERER}
    \end{subfigure}

    \caption{The blue and green boxes highlight two instances of the local exploration paths proposed by Frontier Search and \ours. The former exhibits frequent collisions, leading to inferior exploration efficiency.}
    \label{fig:qualitative_pf}
\end{figure}

The lack of semantic understanding in geometry-based exploration methods like Frontier Search and NBVP clearly impact the amount of road covered and the efficiency of that coverage while running for the same number of steps as \ours. Beyond poor exploratory performance, mapping in outdoor environment, even with ground truth pose and depth values leads to the detection of frontiers and viewpoints that are not traversable, leading to high PF values. Local trajectory visualizations in Fig.~\ref{fig:qualitative_pf} demonstrate the difference in the quality of the exploration paths, at the same locations in Town 5. While using an RGB-only diffusion policy, NoMaD's considerably lower RC and REE values compared to our method highlight the importance of global context and its utilization towards optimizing policy outputs towards actions with higher exploration potential.

The DP-RND baseline was specifically evaluated against our method to highlight the importance of our world model-based VCM over the RND approach to determining dissimilarity between current state and previously observed states. The latent space of the world model used in VCM is better aligned to identify visual similarities/dissimilarities between frames, and with the fast reciprocal matching-based comparison, presents a better metric than RND. The difference in the performance becomes more clear as the towns get larger and lean towards urban settings with greater simulation artifacts such as buildings and four-way intersections.

The qualitative comparison of exploration coverage across three environments in Fig.~\ref{fig:exploration_qualitative} further highlights that \ours achieves superior area coverage across these towns. While learnable diffusion policy-based methods like NoMaD struggle due to a lack of global context and a mechanism to alter the policy accordingly, traditional exploration methods, which are widely used for indoor environments, struggle due to a lack of semantic understanding and mapping artifacts specific to outdoor environments.

\begin{figure}
    \centering
    \includegraphics[width=\linewidth]{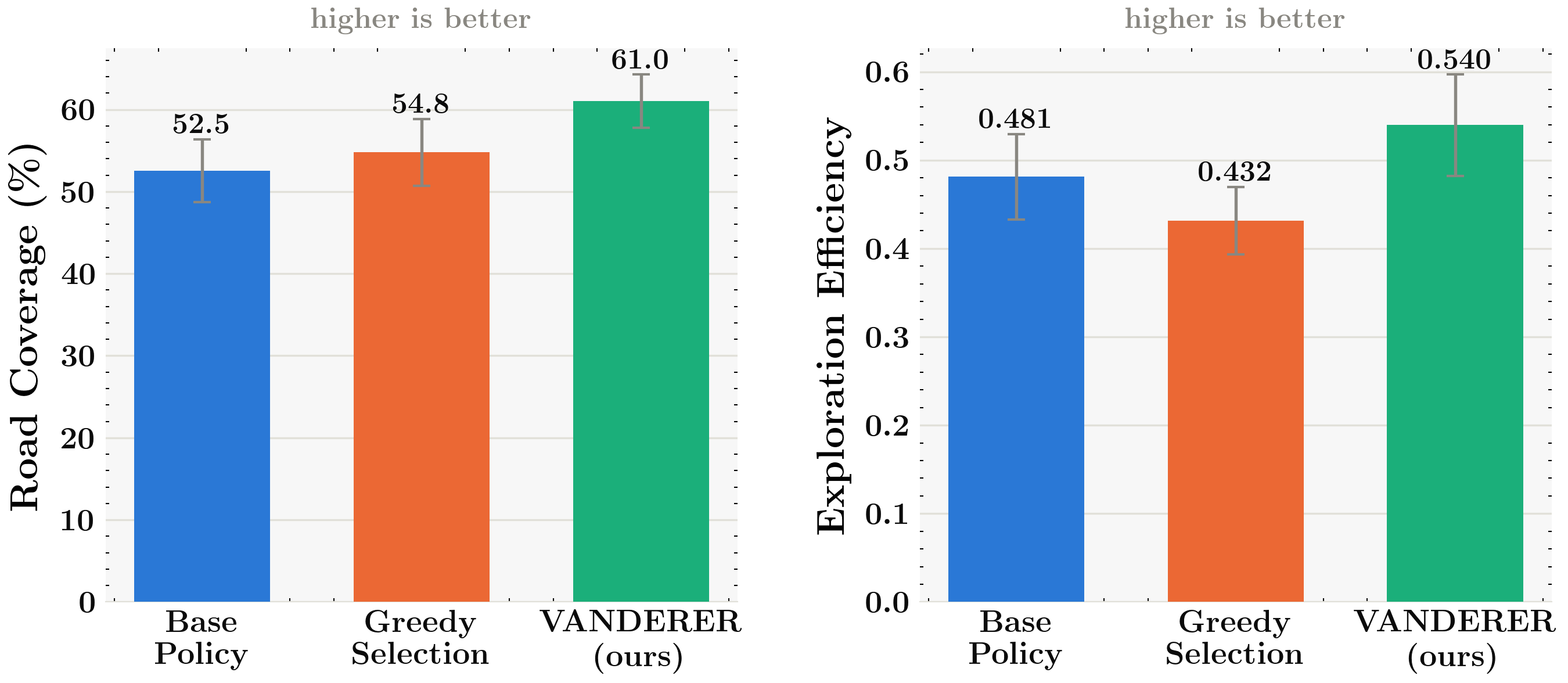}
    \caption{Ablation study of our proposed method. Base Policy evaluates the raw diffusion policy output, while Greedy Selection greedily picks the candidate with the highest VCM score without guidance. Evaluated over all 5 testing environments, \ours shows improved coverage with higher efficiency.}
    \label{fig:guidance_ablations}
\end{figure}

\vspace{1.5mm}
\noindent\textbf{Impact of curiosity-based diffusion guidance.}
A core contribution of \ours is the VCM (visual curiosity-based diffusion guidance). To evaluate its impact on overall exploration performance, we compare our method against two unguided variants: the base diffusion policy and a greedy selection baseline that simply executes the candidate action that leads to the highest curiosity loss. As shown in Fig.~\ref{fig:guidance_ablations}, \ours's diffusion guidance shows improved performance on both the RC and REE metrics, underscoring the importance of our guidance mechanism.

The greedy baseline faces limitations in scenarios where multiple viable exploratory paths exist, e.g., at an intersection where two directions remain unvisited. Because this approach performs independent optimization at each step $t$, it often oscillates between conflicting trajectories at $t$ and $t+1$ rather than committing to a consistent path. Furthermore, the greedy nature of this selection limits performance when immediate candidate actions lead only to previously seen states. In these instances, the lack of stochasticity prevents the agent from discovering unseen locations.
In contrast, \ours's guidance mechanism avoids such issues by updating the noise distribution rather than greedily selecting the best candidate. This helps in retaining a necessary level of randomness when resampling from the updated noise distribution, allowing more robust exploration.

\begin{figure}
    \centering
    \includegraphics[width=1\linewidth]{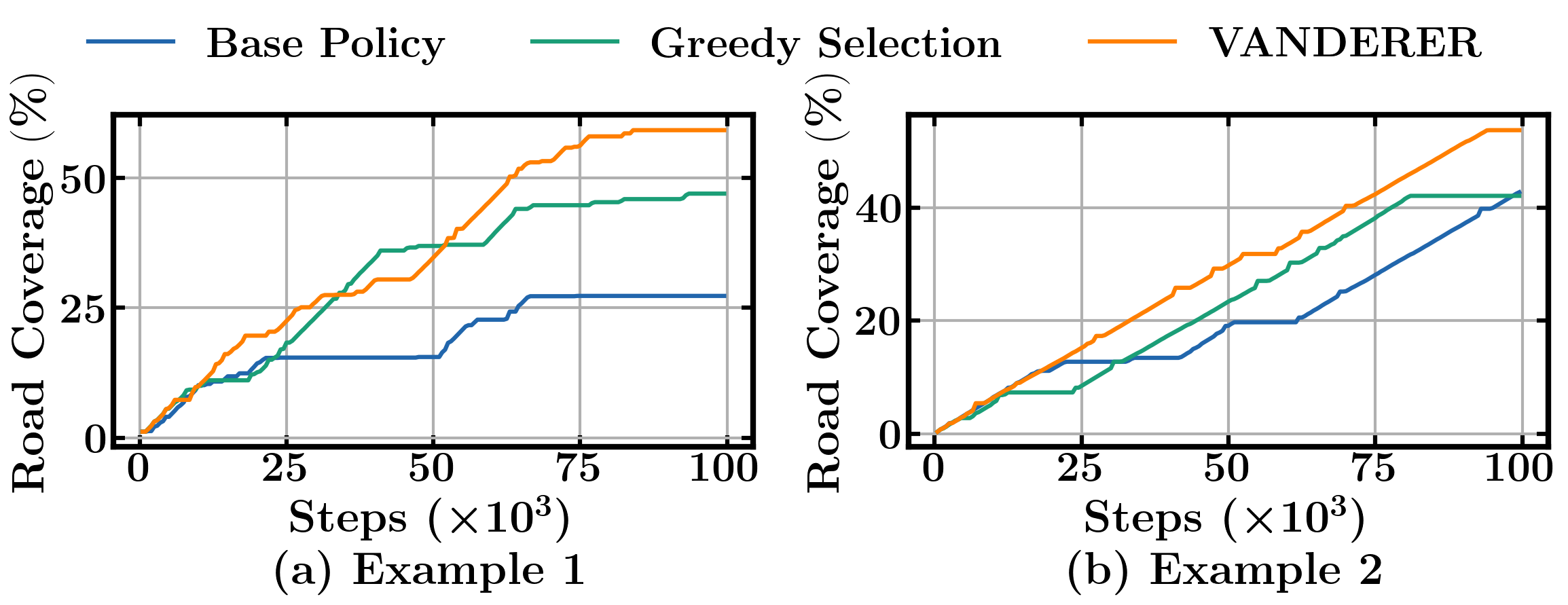}
    \caption{Performance of the Base Policy, Greedy selection (w/o Guidance), and VANDERER (w Guidance) over simulation steps. (a) Initially, most of the environment is unseen, leading to similar performances from all approaches. However, as steps pass, the impact of VCM and guidance starts becoming evident. (b) Greedy selection can sometimes do repeated exploration in loops, resulting in worse performance than the base policy.}
    \label{fig:time_graph}
\end{figure}

\vspace{1.5mm}
\noindent\textbf{Performance with Time.}
The exploration metrics vary with the number of maximum steps taken by the agent. As exploration progresses, the unexplored roads reduce, leading to repetitions which show up as plateaus in the road coverage graph. To understand the behavior of guidance and its importance in \ours, we plot RC for \ours and its ablations (Base Policy and Greedy Selection), against the simulation step count (Fig.~\ref{fig:time_graph}). Initially, when most of the environment is unexplored, the metric grows the same way across all the methods. However, as the simulation progresses, the performance gap widens, and \ours consistently outperforms both baselines.

The comparison between the base policy and greedy selection reveals a notable trend: the greedy selection approach can occasionally repeat road loops. When all paths at a junction have been visited, the greedy agent may get trapped in cyclical trajectories. As evident from the occasional flat lines in Fig.~\ref{fig:time_graph}b, this can cause temporary saturation in RC, negating the advantage of the global context-based selection and leading coverage values similar to the base diffusion policy. This clearly underlines the necessity of the guidance-based strategy to effectively improve the policy's performance as opposed to a greedy selection strategy.

\vspace{1.5mm}
\noindent\textbf{Ablation Studies.}
To evaluate the contribution of the individual components of VANDERER, we conduct ablation studies by systematically removing key architectural elements. Having previously evaluated the impact of the VCM, we now assess the importance of fast reciprocal matching and the diffusion policy. These experiments are conducted in Town 5 of the CARLA dataset, with the metrics in Table \ref{tab:ablations} averaged over three runs with distinct starting poses. The results are discussed below:  \begin{itemize}
    \item \textit{Without Fast Reciprocal Matching}: We replace the fast reciprocal matching module in the VCM with a direct $L_2$-norm comparison between corresponding feature patches of the encoded images. Direct pixel-wise comparison underestimates the similarity between two frames showing similar regions but from different viewpoints because identical features appear at different spatial coordinates. On the contrary our reciprocal matching approach effectively accounts for these geometric variances. This is validated by the performance drop observed in Table \ref{tab:ablations}.
    
    \item \textit{Without Diffusion Policy}: We replace the diffusion policy with a standard Gaussian distribution for sampling action candidates, utilizing CEM for optimization. We maintain identical CEM settings to ensure a fair comparison. The diffusion policy serves as a powerful action prior that significantly accelerates optimization. Ablating this component introduces a fundamental trade-off: a single optimization step (faster computation) yields suboptimal actions and frequent collisions, while increasing optimization iterations renders the method computationally impractical. Consequently, the lack of a strong prior necessitates more intensive optimization, resulting in approximately $55\%$ longer execution time than \ours.
\end{itemize}

\begin{table}[!htbp]
\centering
\caption{Mean performance metrics of ablations across three seeds on Town 5. RC: Road Coverage ($\%$), REE: Road Exploration Efficiency ($m/m$).}
\label{tab:ablations}
\begin{tabular}{l c c c}
\toprule
& {\textbf{RC}} & {\textbf{REE}} & {\textbf{PF}} \\
\midrule
w/o diffusion policy & 20.6 & 0.236 & 7.41\%  \\
w/o fast reciprocal matching & 44.6 & 0.50 & \cellcolor{gray!20}\textbf{0.13\%} \\
Ours & \cellcolor{gray!20}\textbf{59.4} & \cellcolor{gray!20}\textbf{0.67} & 0.25\% \\
\bottomrule
\end{tabular}%
\end{table}

\section{CONCLUSIONS}
Exploring large outdoor environments with limited sensing and traversability constraints remains a significant challenge. To address this, we presented \ours, a framework that couples a diffusion policy with a visual-curiosity-based guidance mechanism to enable efficient exploration using only monocular camera data. Our guidance mechanism samples multiple candidate actions, predicts their outcomes with a world model, and evaluates the novelty of these predictions against previously observed frames. Rather than employing hard selection, we propose a guidance strategy to balance exploration and exploitation, resulting in superior RC with greater exploration efficiency. Through extensive comparisons, we substantiate that visual curiosity is a powerful metric to drive exploration in outdoor environments. Furthermore, our work highlights the potential for pre-trained diffusion models to be guided toward complex exploratory objectives without the need for specialized expert data.


\section*{Acknowledgment}
This work was supported in part by NSF CMMI grant 2208189 and by the New York University Abu Dhabi (NYUAD) Center for Artificial Intelligence and Robotics (CAIR), funded by Tamkeen under the NYUAD Research Institute Award CG010. The funder had no role in study design, data collection and analysis, decision
to publish, or preparation of the manuscript.

\bibliographystyle{elsarticle-num}
\setcitestyle{square}

\bibliography{naviexp}

\end{document}